\documentclass[final,5p,times,twocolumn]{elsarticle}

\usepackage{hyperref}
\usepackage{amssymb}
\usepackage{graphicx}
\usepackage{amsmath}
\usepackage{algorithm}
\usepackage[noend]{algpseudocode}
\usepackage{xcolor}

\newcommand{\Real}{\mathbb{R}}
\usepackage{lineno}
\usepackage{url}

\newtheorem{theorem}{Theorem}
\newtheorem{prop}{Proposition}

\newenvironment{definition}[1][Definition]{\begin{trivlist}
\item[\hskip \labelsep {\bfseries #1}]}{\end{trivlist}}

\newtheorem{col}{Corollary}
\newtheorem{rem}{Remark}

\biboptions{authoryear}

\makeatletter
\def\BState{\State\hskip-\ALG@thistlm}
\makeatother

\begin{document}

\begin{frontmatter}



\title{Stochastic Separation Theorems}


\author[LeicMath]{A.N. Gorban\corref{cor1}}
\ead{ag153@le.ac.uk}
\author[LeicMath,LETI]{I.Y. Tyukin}
\ead{it37@le.ac.uk}

\address[LeicMath]{Department of Mathematics, University of Leicester, Leicester, LE1 7RH, UK}
\address[LETI]{Department of Automation and Control Processes, Saint-Petersburg State Electrotechnical University, Saint-Petersburg, 197376, Russia}

\cortext[cor1]{Corresponding author}

\begin{abstract}

The problem of non-iterative one-shot and non-destructive correction of unavoidable mistakes arises in all Artificial Intelligence applications in the real world.  Its solution requires robust separation of samples with errors from samples where the system works properly. We demonstrate that in (moderately) high dimension this separation could be achieved with probability close to one by linear discriminants.  Based on fundamental properties of measure concentration, we show that for $M<a\exp(b{n})$  random $M$-element sets in $\mathbb{R}^n$ are linearly separable with probability $p$, $p>1-\vartheta$, where $1>\vartheta>0$ is a given small constant. Exact values of $a,b>0$ depend on the probability distribution that determines how the random $M$-element sets are drawn, and on the constant $\vartheta$. These {\em stochastic separation theorems} provide a new instrument
 for the development, analysis, and assessment of machine learning methods and algorithms in high dimension. Theoretical statements are illustrated with numerical examples.
\end{abstract}


\end{frontmatter}

\section{Introduction}

Artificial Intelligence (AI) systems make errors. They should be corrected without damage of existing skills. The  {\em problem of non-destructive correction} arises in many areas of research and development, from AI to mathematical neuroscience, where the reverse engineering of the brain ability to learn on-the-fly remains a great challenge. It is very desirable that the corrector of errors is {\em non-iterative} (one-shot) because iterative re-training of a large system requires much time and resource and cannot be done immediately without impeding activity.

The non-desrructive correction requires separation of the situations (samples) with errors from the samples corresponding to correct behavior by a simple and robust classifier. Linear discriminants introduced by \cite{Fisher1936} are simple, robust, require just the inverse covariance matrix of data, and may be easily modified for assimilation of new data. \cite{Rosenblatt1962} revived the common interest in linear classifiers.
His works sparked intensive scientific debate \citep{Minsky} and gave rise to development of numerous crucial concepts such as e.g. Vapnik-Chervonenkis theory \citep{Vapnik1971}, learnability \citep{Natarajan1989}, and generalization capabilities of neural networks \citep{Vapnik2000}, \citep{Bousquet2002}.
Linear functionals (adaptive summators) are  basic building blocks of significantly more sophisticated AI systems such as e.g. multi-layer perceptrons, \citep{PDP1986}, Convolutional Neural Networks \citep{CNN1995}, \citep{DeepLearning2015} and their derivatives. Much is known about linear functionals as ``stand-alone'' learning machines, including their generalization margins \citep{Frend1999}, \citep{Vapnik2000} and numerous methods for their construction:  linear discriminants  and regression, perceptron learning, and Support Vector Machines \citep{Vapnik1982} among others.

In this work, we demonstrate that in high dimensions and even for exponentially large samples, linear classifiers in their classical Fisher's form are powerful enough to separate errors from correct responses with high probability and to provide efficient solution to the non-destructive corrector problem.
We prove that linear functionals, as learning machines, have surprising and, as far as we are concerned, new peculiar extremal properties: {\it in high dimension, with probability $p>1-\vartheta$ and for $M<a\exp(b{n})$ with $a,b>0$ every point in random i.i.d. drawn $M$-element sets in $\mathbb{R}^n$ is linearly separable from the rest.} Moreover, the separating linear functional can be found explicitly, without iterations. This  property holds for a broad set of relevant distributions, including products of probability measures with bounded support and equidistribution in a unit ball, providing mathematical foundations for one-trial correction of legacy AI systems  (cf. \citep{GorbanRomBurtTyu2016}).

 A problem of data fusion in multiagent systems has clear similarity to the problem of non-destructive correction. According to \cite{Forney2009}, data collected by different agents may not be naively combined due to changes in the context, and special procedures for their assimilation without damage of gained skills are needed. The proven stochastic separation effects can be used to approach this problem.  They also shed light on the possible origins of remarkable selectivity to stimuli observed in-vivo in the real brain \citep{Quiroga2005}.

\section{Preliminaries}

\subsection{Notation}

Throughout the text,  $\Real^n$ is the $n$-dimensional linear real vector space. Unless stated otherwise, symbols $\boldsymbol{x}_i=(x_{i,1},\dots,x_{i,n})$ denote elements of $\Real^n$, and $\big(\boldsymbol{x}_i,\boldsymbol{x}_j\big)=\sum_{k} x_{i,k} x_{j,k}$ is the inner product of $\boldsymbol{x}_i$ and $\boldsymbol{x}_j$ in $\Real^n$. Symbol $\mathbb{B}_n$ stands for the unit ball in $\Real^n$ centered at the origin: $\mathbb{B}_n=\{\boldsymbol{x}\in\Real^n| \ \left(\boldsymbol{x},\boldsymbol{x}\right)\leq 1\}$.

\subsection{Linear Separability of Sets}

\begin{definition}A set $S \subset \mathbb{R}^n$ is {\em linearly separable} if for each $x\in S$ there exists a linear functional $l$ such that $l(x)>l(y)$ for all $y\in S$, $y\neq x$.
\end{definition}
Recall that $x \in \mathbb{R}^n$ is an {\em extreme point} of a convex compact $K$ if there exist no points $y,z\in K$, $y\neq z$ such that $x=(y+z)/2$. The basic examples of linearly separable sets are  extreme points of a convex compacts: vertices of convex polyhedra or points  on the $n$-dimensional sphere. The sets of extreme points of a compact may be not linearly separable  as is demonstrated by simple 2D examples \citep{Simon2011}.

\begin{prop}Every compact linearly separable set $S \subset \mathbb{ R}^n$ is a set of extreme points of a convex compact $K={\rm conv} S$.
\end{prop}
The proof follows immediately from the previous definitions, the Krein-Milman theorem \citep{Simon2011} (its finite-dimensional form was known to Minkovsky) and classical theorems about separation of a point from a convex set.

If $n+1$ points in $\mathbb{R}^n$ do not belong to a hyperplane then they are vertices of a simplex and are, obviously, linearly separable. If the lengths of the edges are bounded then the volume of the simplex decreases with $n$ not slower than $1/n!$. Therefore, we can expect that for a sufficiently regular distribution of points, a random point does not belong to the simplex, and $n+2$ independently chosen random points are also linearly separable, with high probability, which tends to 1 as $1-c/n!$ ($n\to \infty$, $ c>0$).  This fast convergence allows us to hypothesize that even a large random finite set is linearly separable in high dimension with high probability, if the distribution is regular enough.  We prove this statement below for i.i.d. random points from equidistributions in a ball and a cube, and from distributions that are products of measures with bounded support.

\section{Main Results}

\begin{figure}[ht]
\centering\includegraphics[width=0.65\linewidth]{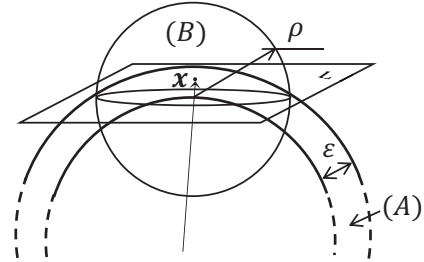}
\caption{Point $\boldsymbol{x}$ belongs to a spherical layer ($A$) of thickness $\varepsilon$. The data are  centralized and the centre of the spheres from $A$ is the origin. Hyperplane $L$ is orthogonal to vector $\boldsymbol{x}$ and tangent to the internal sphere of $A$. $L$ cuts an upper spherical cap from $A$ and separates $\boldsymbol{x}$ from the data points which belong to the external sphere of $A$ but do not belong to that cap.  The cap is included into the upper half of ball ($B$). The centre of $B$ is intersection of the radius $x$ with the internal sphere of the layer $A$.\label{Fig:Balls}}
\end{figure}

Let us start from the equidistribution in the unit ball $\mathbb{B}_n$ in $\mathbb{R}^n$. The probability $p$ that a random point belongs to a layer $\mathbb{B}_n \setminus r\mathbb{ B}_n$ ($0<r<1$) between spheres of radius 1 and of radius $r$ is $p=1-r^n$. Let us take a unit vector $ \boldsymbol{v}$. The probability that the projection of a random vector $\boldsymbol{x}$ on $ \boldsymbol{v}$, $( \boldsymbol{x}, \boldsymbol{v})$, exceeds $r$ can be estimated from above by half of the ratio of volumes of balls of radii $\rho=\sqrt{1-r^2}$ and 1 (see  Fig.~\ref{Fig:Balls} with $\varepsilon=1-r$): $ \mathbf{P}(( \boldsymbol{x}, \boldsymbol{v})>r)\leq 0.5 \rho^{n}$.

\begin{theorem}\label{ball1point}
Let $\{\boldsymbol{x}_1, \ldots , \boldsymbol{x}_M\}$ be a set of $M$  i.i.d. random points  from the equidustribution in the unit ball $\mathbb{B}_n$, $0<r<1$. Then
\begin{equation}\label{Eq:ball1}
\begin{split}
&\mathbf{P}\left(\|\boldsymbol{x}_M\|>r \mbox{ and } \left(\boldsymbol{x}_i,\frac{\boldsymbol{x}_M}{\| \boldsymbol{x}_M\| }\right)<r \mbox{ for all } i\neq M \right)\\
&   \geq 1-r^n-0.5(M-1) \rho^{n};
\end{split}
\end{equation}
\begin{equation}\label{Eq:ballM}
\begin{split}
&\mathbf{P}\left(\|\boldsymbol{x}_j\|>r  \mbox{ and } \left(\boldsymbol{x}_i,\frac{\boldsymbol{x}_j}{\| \boldsymbol{x}_j\|}\right)<r \mbox{ for all } i,j, \, i\neq j\right)\\
&   \geq  1-Mr^n-0.5M(M-1)\rho^{n};
\end{split}
\end{equation}
\begin{equation}\label{Eq:ballMangle}
\begin{split}
&\mathbf{P}\left(\|\boldsymbol{x}_j\|>r  \mbox{ and } \left(\frac{\boldsymbol{x}_i}{\| \boldsymbol{x}_i\|},\frac{\boldsymbol{x}_j}{\| \boldsymbol{x}_j\|}\right)<r \mbox{ for all } i,j, \,i\neq j\right)\\
&  \geq  1-Mr^n-M(M-1)\rho^{n}.
\end{split}
\end{equation}
\end{theorem}

 The proof is based on the independence of random points $\{\boldsymbol{x}_1, \ldots , \boldsymbol{x}_M\}$, on the geometric picture presented in Fig.~\ref{Fig:Balls}, and on an elementary inequality $\mathbf{P}(A_1 \& A_2 \& \ldots \& A_m)\geq 1- \sum_i(1-\mathbf{P}(A_i))$ for any events $A_1, \ldots , A_m$. In Fig.~\ref{Fig:Balls} we should take $\varepsilon=1-r$ and the external radius of the spherical layer $A$ is 1.  \cite{Ball1997} provides more geometric details of concentration of the volume of high-dimensional balls. In (\ref{Eq:ballMangle}) we estimate the probability  that the cosine of the angles between $\boldsymbol{x}_i$ and $\boldsymbol{x}_j$ does not exceed $r$.  \cite{GorbTyuProSof2016} analyzed the asymptotic behavior of these estimations for small $r$. The idea of almost orthogonal bases was introduced by \cite{Kurkova1993} and used efficiently by \cite{Kurkova2007} for estimation of the cardinality of $\varepsilon$-nets in compact convex subsets of Hilbert spaces including the sets of functions computable by perceptrons.

The following corollary gives simple estimates of exponential growth of the maximal possible $M$ for which inequalities (\ref{Eq:ball1}) and (\ref{Eq:ballM}) hold with a given probability value.

 \begin{col}
 Let $\{\boldsymbol{x}_1, \ldots , \boldsymbol{x}_M\}$ be a set of $M$ i.i.d. random points  from the equidustribution in the unit ball $\mathbb{B}_n$ and $0<r,\vartheta<1$. If
\begin{equation}\label{EstimateMball}
M<2({\vartheta-r^n})/{\rho^{n}},
 \end{equation}
 then
 $
 \mathbf{P}((\boldsymbol{x}_i,\boldsymbol{x}_M{)}<r\|\boldsymbol{x}_M\| \mbox{ for all } i=1,\ldots, M-1)>1-\vartheta.
$
 If
 \begin{equation}\label{EstimateM2ball}
M<({r}/{\rho})^n\left(-1+\sqrt{1+{2 \vartheta \rho^n}/{r^{2n}}}\right),
 \end{equation}
  then $\mathbf{P}((\boldsymbol{x}_i,\boldsymbol{x}_j)<r\|\boldsymbol{x}_i\| \mbox{ for all } i,j=1,\ldots, M, \, i\neq j)\geq 1-\vartheta.$

  In particular, if inequality (\ref{EstimateM2ball}) holds then the set $\{\boldsymbol{x}_1, \ldots , \boldsymbol{x}_M\}$ is linearly separable with probability $p>1-\vartheta$.
 \end{col}

 A weaker and simpler estimate (sufficient condition) follows immediately from (\ref{EstimateM2ball}):
 \begin{equation}\label{EstimateM2ballS}
{\vartheta}/ M^2>{r^n+0.5\rho^{n}}.
  \end{equation}
\begin{rem}\label{2,700,000} According to (\ref{EstimateM2ballS}) the pre exponential factor in the estimate for $M^2$ may be chosen as  $\vartheta $, while the exponent depends on $r$ only. For example, for $r=1/\sqrt{2}$ the simple sufficient condition (\ref{EstimateM2ballS}) gives $M^2<\frac{2}{3}\vartheta 2^{n/2}$. For $\vartheta=0.01$ (or specificity 99\%) and $n=100$ we get $M<2,740,000$.

Thus, if we select 2,700,000 i.i.d. points from an equidistribution in a unit ball in $\mathbb{R}^{100} $ then with probability $p>0.99$ {\em all} these points will be vertices of their convex hull.
\end{rem}

Estimates similar to (\ref{Eq:ballMangle}),  (\ref{EstimateM2ball}), and (\ref{EstimateM2ballS}) are useful for the equidistribution of the normalized data on a unit sphere too. This is because they not only establish the fact of separability but also specify separation margins.

Consider a product distribution in an $n$-dimensional unit cube. Let the coordinates of a random point, $X_1, \ldots, X_n$ ($0 \leq X_i \leq 1$) be independent random variables with expectations $\overline{X}_i$ and variances $\sigma_i^2>\sigma_0^2>0$. Let $\overline{\boldsymbol{x}}$ be a vector with coordinates $\overline{X}_i$. For large $n$, this distribution is concentrated in a relatively small vicinity of a sphere with an {\em arbitrary} centre $\boldsymbol{c}$ with coordinates $c_i$ and radius $R$, where

\begin{equation}\label{ConcRad}
R^2=\mathbf{E}\left(\sum_i (X_i-c_i)^2\right)=\sum_i \sigma_i^2  + \|\overline{\boldsymbol{x}}-\boldsymbol{c}\|^2.
\end{equation}
Concentration near the spheres with different centres implies concentration in the vicinity of their intersection (an example of the `waist concentration' \citep{Gromov2003}). The vicinity of the spheres, where the distribution is concentrated, can be estimated by the Hoeffding inequality \citep{Hoeffding1963}. Let  $Y_1, \ldots, Y_n$ be independent bounded random variables: $0 \leq Y_i \leq 1$. The empirical mean of these variables is defined as $\overline Y = \frac{1}{n}(Y_1 + \cdots + Y_n)$. Then

\begin{equation}
\begin{split}
&\mathbf{P} \left(\overline Y - \mathbf{E}\left [\overline Y \right] \geq t \right) \leq \exp \left(-2n t^2 \right); \\
&\mathbf{P} \left(\left |\overline Y - \mathbf{E}\left [\overline Y \right] \right | \geq t \right) \leq 2\exp \left(-2nt^2 \right).
\end{split}
\end{equation}

 Let us take $Y_i=(X_i-c_i)^2$. Consider the centres located in the cube, $0\leq c_i \leq 1$. Then $0\leq Y_i \leq 1$ and $\mathbf{E}\left [\overline Y \right]=\frac{1}{n}R^2$. In particular, if $c_i=\overline{X}_i$ then $\mathbf{E}\left [\overline Y \right]=\frac{1}{n}R_0^2$ (the minimal possible value), where $R_0^2=\sum_i \sigma_i^2\geq n\sigma_0^2$. In general, $n\sigma_0^2\leq R^2 \leq n$.

 With probability $p>1- 2\exp \left(-2nt^2 \right)$ a random point $\boldsymbol{x}$ belongs to the spherical layer ($\delta={nt}/{R_0^2}$, $t=\delta R_0^2/n$):
 \begin{equation}\label{sphericLayer}
1-\delta \leq {\|\boldsymbol{x}-\overline{\boldsymbol{x}}\|^2}/{R_0^2}\leq 1+\delta.
\end{equation}

Consider $M$ i.i.d. points $\{\boldsymbol{x}_1, \ldots , \boldsymbol{x}_M\}$ from the product distribution. With probability $p>1- 2M\exp \left(-2nt^2 \right)$ they all belong to the spherical layer (\ref{sphericLayer}). Therefore, with this probability we return to the situation presented in Fig.~\ref{Fig:Balls}  with internal radius $\sqrt{1-\delta}R_0$ and external radius $\sqrt{1+\delta}R_0$. The difference from the equidistribution in the ball is that the volume of the ball is concentrated near the external sphere, while the distribution in the layer (\ref{sphericLayer}) is concentrated around the sphere $\|\boldsymbol{x}-\overline{\boldsymbol{x}}\|^2=R_0^2$.

The radius of ball $B$ is defined by $\rho^2=(1+\delta)R_0^2-(1-\delta)R_0^2=2\delta R_0^2$.
The concentration radius (\ref{ConcRad}) for the spheres concentric with the ball $B$ (Fig.~\ref{Fig:Balls}) is defined by $R^2=R_0^2+(1-\delta)R_0^2=(2-\delta)R_0^2$.
Therefore,  a random point does not belong to the ball $B$ with probability $p>1- \exp \left(-2n\tau^2 \right)$,  where
$\tau=\frac{1}{n}(R^2-\rho^2)=\frac{1}{n}(2-3 \delta)R^2_0$.
 Thus,  we get the following statement.

\begin{theorem}\label{cube} Let  $\{\boldsymbol{x}_1, \ldots , \boldsymbol{x}_M\}$ be i.i.d. random points  from the product distribution in a unit cube, $0< \delta <2/3$. Then
\begin{equation}\label{Eq:cube1}
\begin{split}
&\mathbf{P}\left(1-\delta  \leq \frac{\|\boldsymbol{x}_j-\overline{\boldsymbol{x}}\|^2}{R^2_0}\leq 1+\delta \mbox{ and } \right. \\
& \left.
\left(\frac{\boldsymbol{x}_i-\overline{\boldsymbol{x}}}{R_0},\frac{\boldsymbol{x}_M-\overline{\boldsymbol{x}}}{\| \boldsymbol{x}_M-\overline{\boldsymbol{x}}\| }\right)<\sqrt{1-\delta} \mbox{ for all } i,j, \, i\neq M \right)\\
&\geq 1- 2M\exp \left(-2\delta^2 R_0^4/n \right) -(M-1)\exp \left(-2R_0^4(2-3 \delta)^2/n\right);
\end{split}
\end{equation}
\begin{equation}\label{Eq:cube2}
\begin{split}
&\mathbf{P}\left(1-\delta  \leq \frac{\|\boldsymbol{x}_j-\overline{\boldsymbol{x}}\|^2}{R^2_0}\leq 1+\delta \mbox{ and } \right. \\
&\left.
\left(\frac{\boldsymbol{x}_i-\overline{\boldsymbol{x}}}{R_0},\frac{\boldsymbol{x}_j-\overline{\boldsymbol{x}}}{\| \boldsymbol{x}_j-\overline{\boldsymbol{x}}\| }\right)<\sqrt{1-\delta} \mbox{ for all } i,j, \, i\neq j \right)\\
&\geq 1- 2M\exp \left(-2\delta^2 R_0^4/n \right) -M(M-1)\exp \left(-2R_0^4(2-3 \delta)^2/n\right)
\end{split}
\end{equation}
\end{theorem}
When the value of delta is chosen as $\delta=0.5$ and $R_0$ is replaced with its estimate from below, $R^2_0\geq n \sigma_0^2$, inequalities  (\ref{Eq:cube1})  and  (\ref{Eq:cube2}) result in the following  estimates:
\begin{equation}\label{Eq:cube3}
\begin{split}
&\mathbf{P}\left(\frac{1}{2}  \leq \frac{\|\boldsymbol{x}_j-\overline{\boldsymbol{x}}\|^2}{R^2_0}\leq \frac{3}{2} \mbox{ and } \right. \\
& \left.
\left(\frac{\boldsymbol{x}_i-\overline{\boldsymbol{x}}}{R_0},\frac{\boldsymbol{x}_M-\overline{\boldsymbol{x}}}{\| \boldsymbol{x}_M-\overline{\boldsymbol{x}}\| }\right)<\sqrt{1-\delta} \mbox{ for all } i,j, \, i\neq M \right)\\
&\geq  1-3M\exp\left(-0.5 n\sigma_0^4 \right);
\end{split}
\end{equation}
\begin{equation}\label{Eq:cube4}
\begin{split}
&\mathbf{P}\left(\frac{1}{2}  \leq \frac{\|\boldsymbol{x}_j-\overline{\boldsymbol{x}}\|^2}{R^2_0}\leq\frac{3}{2} \mbox{ and } \right. \\
&\left.
\left(\frac{\boldsymbol{x}_i-\overline{\boldsymbol{x}}}{R_0},\frac{\boldsymbol{x}_j-\overline{\boldsymbol{x}}}{\| \boldsymbol{x}_j-\overline{\boldsymbol{x}}\| }\right)<\sqrt{1-\delta} \mbox{ for all } i,j, \, i\neq j \right)\\
&\geq 1-M(M+1)\exp\left(-0.5 n\sigma_0^4 \right).
\end{split}
\end{equation}

\begin{col}
 Let  $\{\boldsymbol{x}_1, \ldots , \boldsymbol{x}_M\}$ be i.i.d. random points  from the product distribution in a unit cube and $0<\vartheta<1$. If
  \begin{equation}\label{SampleSizeCube1}
 M<\frac{1}{3}\vartheta \exp\left(0.5 n\sigma_0^4 \right),
 \end{equation}
  then with probability $p>1-\vartheta$
\[
\begin{split}
&  0.5\leq \frac{\|\boldsymbol{x}_j-\overline{\boldsymbol{x}}\|^2}{R_0^2} \leq 1.5 \mbox{ and }
\left(\frac{\boldsymbol{x}_i-\overline{\boldsymbol{x}}}{R_0},\frac{\boldsymbol{x}_M-\overline{\boldsymbol{x}}}{\| \boldsymbol{x}_M-\overline{\boldsymbol{x}}\|}\right)<\frac{\sqrt{2}}{2} \\
&\mbox{  for all } i,j, i \neq M.
\end{split}
\]
 If
  \begin{equation}\label{SampleSizeCube2}
(M+1)^2<\frac{1}{3}\vartheta \exp\left(0.5 n\sigma_0^4 \right),
 \end{equation}
   then with probability $p>1-\vartheta$

\[
\begin{split}
&0.5\leq \frac{\|\boldsymbol{x}_j-\overline{\boldsymbol{x}}\|^2}{R_0^2} \leq 1.5 \mbox{ and }
\left(\frac{\boldsymbol{x}_i-\overline{\boldsymbol{x}}}{R_0},\frac{\boldsymbol{x}_j-\overline{\boldsymbol{x}}}{\| \boldsymbol{x}_j-\overline{\boldsymbol{x}}\|}\right)<\frac{\sqrt{2}}{2} \\
& \mbox{ for all }i,j, i\neq j.
\end{split}
\]
In particular, if inequality (\ref{SampleSizeCube2}) holds then the set   $\{\boldsymbol{x}_1, \ldots , \boldsymbol{x}_M\}$ is linearly separable with probability $p>1-\vartheta$.
   \end{col}

\begin{figure}
\centering
\includegraphics[width=0.8\linewidth]{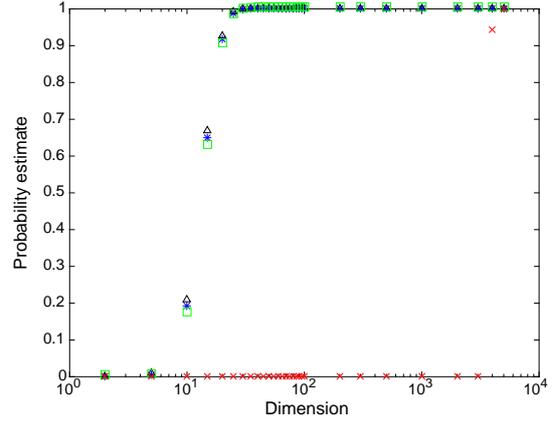}
\caption{Estimates of probabilities that a random point in a sample of $20 000$ points i.i.d. drawn from an equidistribution in the unit cube $[0,1]^n$ in $\Real^n$ is separable from the remaining points in the sample as a function of dimension $n$.   Blue stars, black triangles, and green squares, are the means, maxima, and minima of $N_{-}/N$ over all $100$ samples for each value of $n$. Red crosses show estimates (\ref{Eq:cube3}).}\label{fig:experiments_cube}
\end{figure}

The estimates (\ref{SampleSizeCube1}), (\ref{SampleSizeCube2}) are far from being optimal and can be improved. The main message here is their exponential dependence on $n$: the upper boundary of $M$ can grow with $n$ exponentially. Numerical experiments show that the equidistribution in cube is not worse, from the practical point of view, than the uniform distribution in a ball. To illustrate this, we empirically assessed linear separability of samples drawn from equidistributions in the unit $n$-cubes. For selected values of $n$ from the set $\{1,\dots,5000\}$ we generated $100$ samples $S$ of $M=20 000$ random points from $[0,1]^n$. For each sample, a sub-sample $\underline{S}\subset S$ of $N=4000$ points was randomly chosen, and for each point $\boldsymbol{x}_i$ in this sub-sample linear functionals $l(\boldsymbol{x})=\left(\boldsymbol{x}_i-\boldsymbol{\bar{x}},\boldsymbol{x}-\boldsymbol{\bar{x}}\right)-\|\boldsymbol{x}_i-\boldsymbol{\bar{x}}\|^{2}$ were constructed. Sings of $l(\boldsymbol{x}_j)$, $\boldsymbol{x}_j\in S$, $\boldsymbol{x}_j\neq \boldsymbol{x}_i$ were calculated, and the numbers $N_{-}$ of instances  when $l(\boldsymbol{x}_j)< 0$ where recorded. Empirical frequencies $N_{-}/N$ were then derived. Outcomes of this experiment are summarized in Fig. \ref{fig:experiments_cube}. These experiments demonstrate that the probability that a randomly selected point in a sample is linearly separable from the rest could be significantly higher than the simple exponential estimates provided. This, however, is not surprising as the estimates are based on the values of means and variances, and do not take into account other quantitative properties of the sample distribution.

In general position, a set of $n$ points in $\mathbb{R}^{n-1}$ is linearly separable. Therefore, if $n-1$ or less points from $\mathcal{M}=\{\boldsymbol{x}_1, \ldots , \boldsymbol{x}_{M-1}\}$ are not separated from $\boldsymbol{x}$ by the hyperplane $L$ (Fig.~\ref{Fig:Balls}) then they can be separated by an additional hyperplane orthogonal to $L$. This means that  $\boldsymbol{x}$  can be separated from the whole $\mathcal{M}$ by a
conjunction of two linear inequalities,
$(\bullet , \boldsymbol{x}/\|\boldsymbol{x}\|) >r \, \& \, (\bullet , \boldsymbol{y}) > q$, for some $0<r<0$, $q>0$, and $\boldsymbol{y}$, $(\boldsymbol{y},\boldsymbol{x})=0$. This system can be considered as a cascade of two independent neurons \citep{GorbanRomBurtTyu2016}.  The probability of such a {\em two-neuron separability} is higher than of linear separability. (Compare inequality (\ref{Eq:ball1n}) in the following theorem to (\ref{Eq:ball1}).)

\begin{theorem}\label{CascadeSep}
Let $S=\{\boldsymbol{x}_1, \ldots , \boldsymbol{x}_M\}$ be a set of $M$  i.i.d. random points  from the equidustribution in the unit ball $\mathbb{B}_n$, $0<r<1$. Then
\begin{equation}\label{Eq:ball1n}
\begin{split}
&\mathbf{P} \left(\|\boldsymbol{x}_M\|>r \, \&\,  \left(\boldsymbol{x}_i,\frac{\boldsymbol{x}_M}{\| \boldsymbol{x}_M\| }\right)<r
 \mbox{ for at least } M\!-\!n \mbox{ points }  \boldsymbol{x}_i\in S\right)  \\
& \geq (1-r^n)(1-0.5\rho^{n})^{M-1} \\
&\times  \left(1-\frac{1}{n!}\left(\frac{0.5(M-n)\rho^{n}}{1-0.5\rho^{n}}\right)^n\right) \exp{\left[\frac{0.5(M-n)\rho^{n}}{1-0.5\rho^{n}}\right]},
\end{split}
\end{equation}
\end{theorem}
where $\rho=\sqrt{1-r^2}$.

For $r=1/\sqrt{2}$, $n=100$, and $M=2,74\cdot 10^6$, (\ref{Eq:ball1n}) gives:  $\mathbf{P} \left(\|\boldsymbol{x}_M\|>r \& \left(\boldsymbol{x}_i,\frac{\boldsymbol{x}_M}{\| \boldsymbol{x}_M\| }\right)<r
 \mbox{ for at least } M\!-\!n \ \ \boldsymbol{x}_i\ \in S\right) \geq 1-\theta$ with $\theta<5\cdot 10^{-14}$. The probability stays close to $1$ for much larger values of  $M$, as setting $M=7\cdot 10^{16}$ results in the estimate: $\mathbf{P} \left(\|\boldsymbol{x}_M\|>r \& \left(\boldsymbol{x}_i,\frac{\boldsymbol{x}_M}{\| \boldsymbol{x}_M\| }\right)<r
 \mbox{ for at least } M\!-\!n \ \ \boldsymbol{x}_i\in S\right) \geq 1-\theta$ with $\theta<5\cdot 10^{-9}$.

\section{Conclusion}

Classical measure concentration theorems state that random points are concentrated in a thin layer near a surface (a sphere, an average or median level set of energy or another function, etc.).
The {\em stochastic separation theorems} describe  thin structure of these thin layers: the random  points are not only concentrated in a thin layer but are all linearly separable from the rest of the set even for exponentially large random sets. The estimates are produced for two classes of distributions in high dimension: for equidistributions in balls or ellipsoids or for the product distributions with compact support (i.e. for the case when coordinates are bounded independent random variables).  Numerous generalisations are possible, for example:
\begin{itemize}
\item Relax the requirement of independent coordinates in Theorem~\ref{cube} to that of weakly dependent vector-valued variables;
\item Instead of equidistributions, consider distributions with strongly log-concave probability  densities;
\item Use various simple and robust nonlinear classifiers like small neural cascades (compare to Theorem~\ref{CascadeSep}), algebraic classifiers and other families. For these generalisations, the VC dimension is expected to play the  role similar to  dimension $n$ in Theorems \ref{ball1point} and \ref{cube}.
\end{itemize}

Stochastic separation Theorems 1--3 are important for synthesis and one-shot correction of AI systems. For example, inequalities (\ref{ball1point}) and (\ref{Eq:cube1}) evaluate the probability that a randomly selected point $\boldsymbol{x}_M$ is linearly separable from all other $M-1$ points by the linear functional $l(\boldsymbol{x})=(\boldsymbol{x},\boldsymbol{x}_M-\overline{\boldsymbol{x}})$. This separation is sufficient to correct a mistake of a legacy AI system without any re-learning and modification of existing skills \citep{GorbanRomBurtTyu2016}. Such measure concentration effects reveal the hidden geometric background of the reported success of randomized neural networks models \citep{ScardapaneWang2017}.

Stochastic separation theorems can simplify high-dimensional data analysis and generate the 'blessing of dimensionality' \citep{GorbanTyuRom2016}.  For example, according to (\ref{EstimateM2ballS}), in a dataset with 100 attributes and $M<2.7\cdot 10^6$ samples we should not be surprised to observe the  linear separability of {\em each} sample from the rest of the database by the inequalities $\langle \boldsymbol{x}_i,\boldsymbol{x}_j\rangle<\sqrt{\frac{1}{2}\langle\boldsymbol{x}_i,\boldsymbol{x}_i\rangle}$ ($i\neq j$) in the Mahalanobis inner product $\langle \boldsymbol{x}, \boldsymbol{y}\rangle =(\boldsymbol{x}, S^{-1}\boldsymbol{y})$, where $S$ is the empirical covariance matrix. The Mahalanobis inner product is used for `whitening', i.e. for transformation of the data cloud into the spherical form. Of course, these attributes should not be highly correlated and the empiric covariance matrix  should be invertible.

We analysed separation of random points from random sets. This is the problem of single {\em correction} of a legacy AI system. The question of generalisability of this correction is of  great practical importance. It leads to a problem of {\em separation of two random sets}. A simple series of generalisations can be immediately produced from Theorems 1-3 for separation of an $M$-element random set $S=\{\boldsymbol{x}_1, \ldots , \boldsymbol{x}_M\} \subset \mathbb{R}^{n}$ from a $k$-element one  $\{\boldsymbol{y}_1, \ldots , \boldsymbol{y}_k\}$ for $k<n$. For this purpose, we can consider a linear space $E=\mathrm{span}\{\boldsymbol{y}_i-\boldsymbol{y}_1 \, | \, i=2,\ldots , k\}$ and study separation of a point from an $M$-element set in the projection onto the quotient space $\mathbb{R}^{n}/E$. If  ${\boldsymbol{y}_1,\ldots,\boldsymbol{y}_k}$ are independent  then separation would likely be limited to sets of small cardinality $k<n$. If, in contrast,   ${\boldsymbol{y}_1,\ldots,\boldsymbol{y}_k}$ are pair-wise positively correlated then we can expect that  a single functional would separate them from $S$, with reasonable probability even for some $k\geq n$. This naturally gives rise to generalization of “corrections”.

The reported extreme separation capabilities of linear functionals offer new insights into the Grandmother cell or concept cell phenomena that are broadly reported in neuroscience \citep{Quiroga2005}, \citep{Quiroga2009}. The essence of the phenomenon is that some neurons in the  human brain respond unexpectedly selective  to particular persons or objects. Strikingly, not only is the brain able to respond selectively to ``rare'' individual stimuli but also such selectivity can be learnt very rapidly from a limited number of experiences \citep{Quiroga2015}. The question is: Why small ensembles of neurons may deliver such a sophisticated functionality reliably? Stochastic separation Theorems 1-3 provide a possible answer. If we accept that a) linear functionals followed by nonlinear threshold-modulated response as phenomenological models of cells whose activity was measured, b) the number of inputs converging to these cells is large enough, and c) they are statistically independent, then extreme selectivity of responses of such models follows immediately from Theorem 2.

%In this projection, the set $\{\boldsymbol{y}_1, \ldots , \boldsymbol{y}_k\}$ maps into one point and we return to the problem of separation of  one random point from a random $M$-element set in $(n-k)$-dimensional space. The concentration radius $R$ scales at such a projection as $\sqrt{(n-k)/n}$.

\section*{Acknowledgement}
Authors are grateful to M. Gromov and S. Utev for seminal questions and comments. The work was partially supported by Innovate UK (KTP009890 and KTP010522).

\end{document}